\documentclass{article}

\usepackage{PRIMEarxiv}

\usepackage[utf8]{inputenc} 
\usepackage[T1]{fontenc}    
\usepackage{hyperref}       
\usepackage{url}            
\usepackage{booktabs}       
\usepackage{amsfonts}       
\usepackage{nicefrac}       
\usepackage{microtype}      
\usepackage{lipsum}
\usepackage{fancyhdr}       
\usepackage{graphicx}       
\graphicspath{{media/}}     
\usepackage{indentfirst} 
\title{ Video Text Tracking for Dense and Small Text Based on PP-YOLOE-R and Sort Algorithm

\author{
  Hongen Liu \\
  y2998388548@163.com} 
}

\begin{document}
\maketitle{}
\begin{abstract}
Although end-to-end video text spotting methods based on Transformer can model long-range dependencies and simplify the train process, it will lead to large computation cost with the increase of the frame size in the input video. Therefore, considering the resolution of ICDAR 2023 DSText is $1080\times1920$ and slicing the video frame into several areas will destroy the spatial correlation of text, we divided the small and dense text spotting into two tasks, text detection and tracking. For text detection, we adopt the PP-YOLOE-R which is proven effective in small object detection as our detection model. For text detection, we use the sort algorithm for high inference speed. Experiments on DSText dataset demonstrate that our method is competitive on small and dense text spotting. 
\end{abstract}


\section{Introduction}

\par While existing video text spotting \cite{textpsottng} including ICDAR 2015 \cite{icdar2105}, YouTube Video Text \cite{youtube}, RoadText-1K \cite{roadtext}, BOV Text \cite{bovtext} focus on common text case, less attention was paid on dense and small text spotting. However, the dense and small text is crucial for computer vision perception in the distance and occupied a high percentage in open world scenarios. Comparable to common text case, the dense and small text is more susceptible to the background and external interference such as motion blur, out of focus, and artifacts issues. 
\par Existing video text spotting methods can be divided into two paradigms, three-staged pipeline \cite{bovtext,three_1,three_2} and end-to-end pipeline \cite{wu2022transdetr}. While the three-staged pipeline requires to train the detection network, recognition network and tracking network individually,  The end-to-end pipelines considers the video text spotting task as a direct long-range temporal modeling problem, detecting, recognizing and tracking the text end-to-end. Nevertheless, the small and text videos usually have high resolution such as $1080\times1920$, which will lead to expensive computational cost when transformer pipeline is used. Slicing the whole images into several areas and predicting each areas individually may destroy the spatial correlation. Restricted to the computational resources, we adopt the three-staged pipeline.
\par Although the semantic information and character level annotation are useful in the common text case, we believe that the above information is less effective to dense and small text, but brings extra computation overhead. Since there is severe image degradation on dense and small text instance, the semantic and character feature are significantly weakened. Therefore, we view the video text spotting for dense and small text as the small object detection problem.Here, we only participated in the Task One, video text tracking in the competition and designed a two-stage network based on the PP-YOLOE-R \cite{ppyoloe_r} and sort \cite{Bewley2016_sort} algorithm.The PP-YOLOE-R is an efficient anchor-free rotated detector proposed by Baidu and  achieves 78.14 mAP in large version on DOTA 1.0 dataset, which is a famous small object detection dataset in aerial image. And the sort algorithm proposed by Bewley, is a simple and fast method in multiple object tracking.

\section{Method}
\label{sec:headings}
The proposed method tackles the small and dense text detection as the small object detection problem and ignores the textual characteristic. The overall architecture is depicted in Figure 1, which contains two main components: the PP-YOLOE-R network and Sort algorithm. First of all, we divided the video into a series of image frames. Then, PP-YOLOE-R network uses the individual image frame as the input and outputs the rotated detection box in each frame. After obtaining the detection results of each video, we adopt the Sort algorithm to assign identical traceID for the same small and dense text among frames.

\begin{figure}
  \centering
 \includegraphics[width=6.5in]{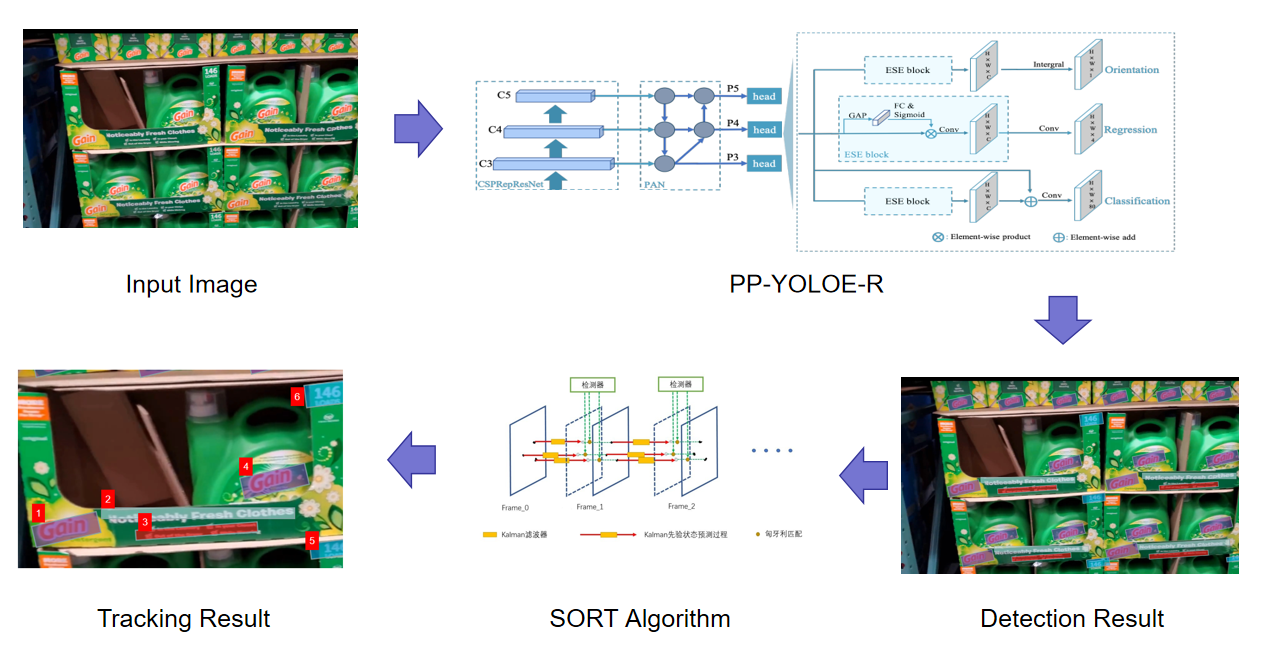}
  \caption{The overall architecture of our method.}
  \label{fig:fig1}
\end{figure}

\section{Experiments}
\subsection{Datasets}
\par \textbf{ICDAR2023 DSText} is proposed in the ICDAR2023 Video Text Reading Competition for Dense and Small Text, which contains 50 videos for training and 50 videos for testing. These videos are harvested from activity, driving, game, sports, indoor street and outdoor street scenarios, and each text is labeled as quadrangle with 4 vertexes in word-level. Transcriptions and trace id are provided.
\subsection{Implementation Details}
All the experiments are conducted on Paddle with Tesla V100 32G GPUs provided by Baidu AI studio. We use the PP-YOLOE-R \cite{ppyoloe_r} as our basic network. The model is trained on the ICDAR2023 DSText. We divided the videos into a series of static image frame as the input of our model and adopt the Momentum as the optimizer with the initial rate of 8e-3 and momentum of 0.9. The PP-YOLOE-R was trained with cosine decay schedulers for 36 epochs and linear warmUp was adopted at first 1000 steps to avoid the instability at the initial training stage. The batch size is set to be 5 during training and input size is set to $900\times1600$. Following the PP-YOLOE-R, the data augmentation includes random flip, random rotate at 0°, 90° , 180°, -90°, random rotates ranging from 30° to 90°. The inference are tested with a batch size of 1 on a V100 GPU.
\subsection{ Visualization of Small and Dense Text Tracking}
\begin{figure*}
  \centering
   \includegraphics[width=6.45in]{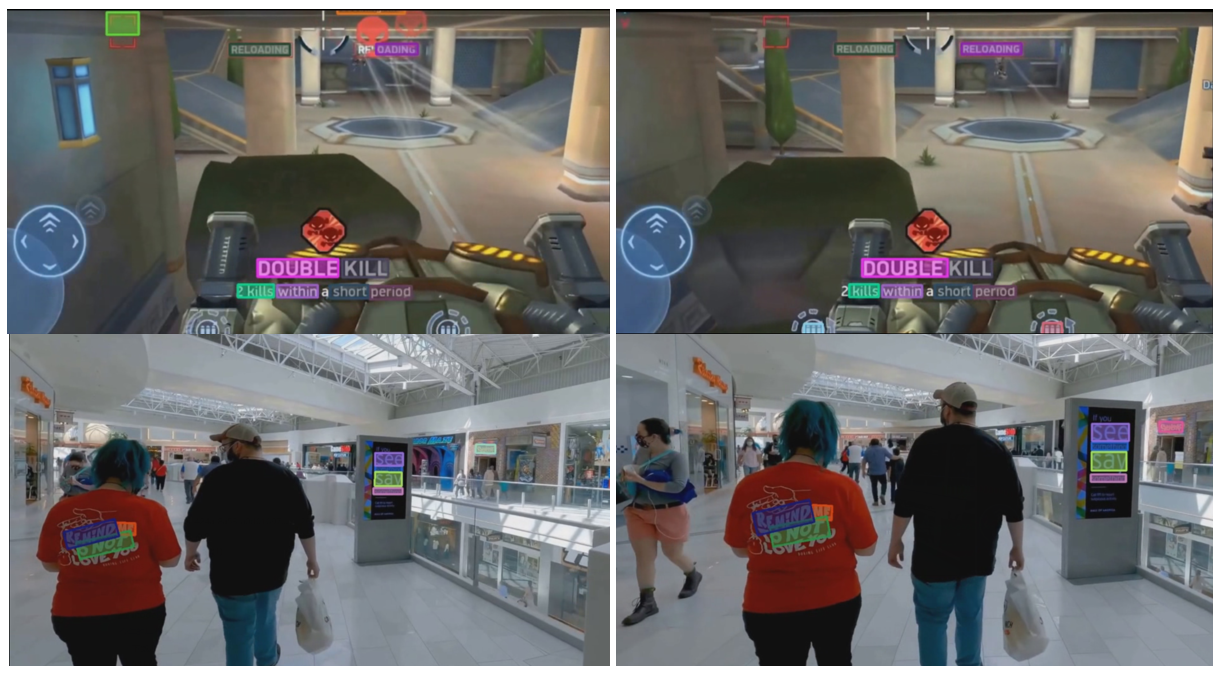}
  \caption{The visualization of detection and tracking results in game and activity  scenarios}\label{vis1}
\end{figure*}\begin{figure*}
  \centering
   \includegraphics[width=6.45in]{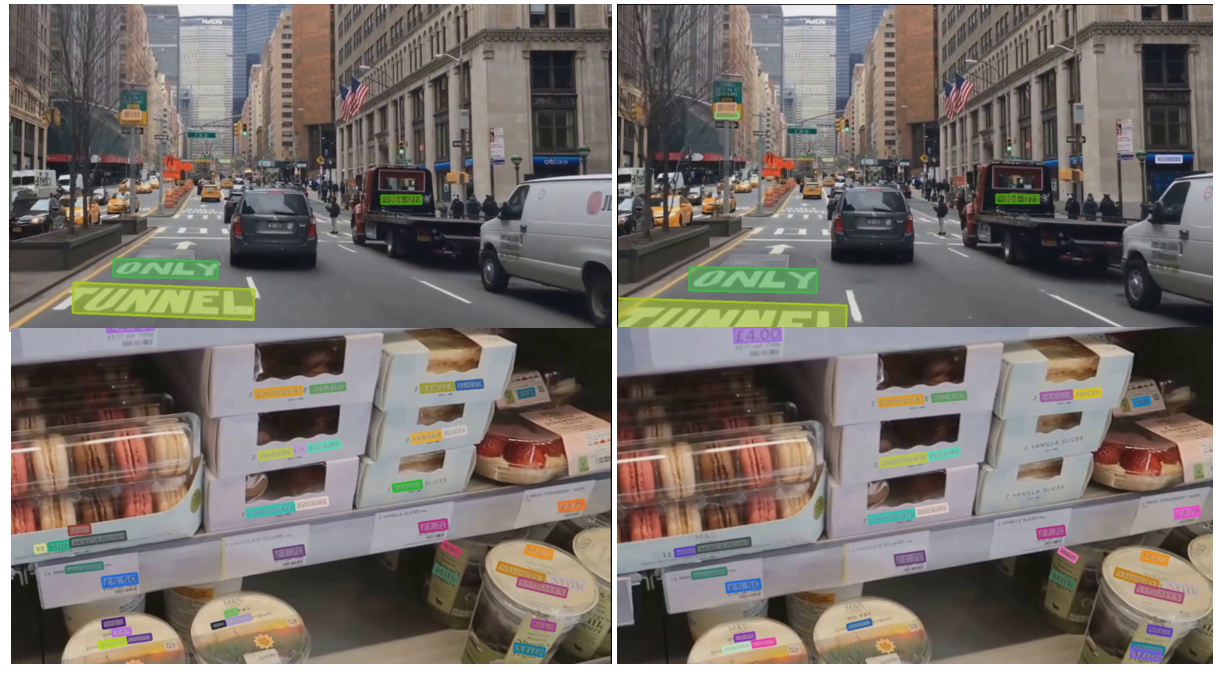}
  \caption{The visualization of detection and tracking results in driving and indoor street scenarios}\label{vis2}
\end{figure*}
\par We visualizes the detection and tracking results of our method, which is shown in Fig. 2 and Fig. 3. We cropped and enlarge the local areas due to high resolution of original video. The same color denotes the same trace id among frames. It can be seen that our method achieves competitive performance in different scenarios.

\section{Conclusion}
In this paper, we divided the small and dense text tracking into two subtasks, text detection and tracking. While conventional method considered the small and dense text as text detection problem, we view the task as small object detection problem. Thus, we use the start of arts model in small object detection, PP-YOLOE-R as our base model and adopt the sort algorithm to track the localized text for high inference speed. The visualization of tracking results on different scenarios validate the effectiveness of our method.

\bibliographystyle{unsrt}  
\bibliography{references}

\end{document}